\documentclass{article}

\usepackage[nonatbib, final]{infocog_neurips_2022}
\usepackage[numbers]{natbib}

\usepackage{graphicx}

\usepackage{caption, subcaption}

\makeatletter
\newcommand{\nocaption}{\addtocounter{\@captype}{-1}}
\makeatother

\usepackage[utf8]{inputenc} 
\usepackage[T1]{fontenc}    
\usepackage{hyperref}       
\usepackage{url}            
\usepackage{booktabs}       
\usepackage{amsfonts}       
\usepackage{nicefrac}       
\usepackage{microtype}      
\usepackage{xcolor}         

\title{Higher-order mutual information reveals synergistic sub-networks for multi-neuron importance}
\author{%
Kenzo Clauw$^{1}$ \quad Daniele Marinazzo$^{1}$ \quad Sebastiano Stramaglia$^2$ \\
$^1$Department of Data Analysis, Ghent University \quad $^2$Physics Department, University of Bari\\
\texttt{kenzoclauw@gmail.com}\\
\texttt{daniele.marinazzo@ugent.be}\\
\texttt{sebastiano.stramaglia@uniba.it}
}

\begin{document}

\maketitle

\begin{abstract}
  Quantifying which neurons are important with respect to the classification decision of a trained neural network is essential for understanding their inner workings. Previous work primarily attributed importance to individual neurons. In this work, we study which groups of neurons contain synergistic or redundant information using a multivariate expansion of the mutual information (O-information). We observe that the first layer is dominated by redundancy, suggesting general shared features (i.e. detecting edges), while the last layer is dominated by synergy, indicating local class-specific features (i.e. concepts). Finally, we show that the O-information can be used for multi-neuron importance. This can be demonstrated by re-training a synergistic sub-network, which results in a minimal change in performance. These results suggest that our method can be used for pruning and unsupervised representation learning. 
\end{abstract}

\section{Introduction}
Methods for understanding deep neural networks (DNNs) often focus on analysing properties of individual neurons to explain their generalization performance. One approach is to quantify the importance of neurons for the classification task based on their selectivity (i.e. neurons firing only for a specific class)\cite{morcos_importance_2018}. Various studies have shown that DNNs might not rely on these selective neurons for generalization \cite{liu_understanding_2018}\cite{matthew_l_leavitt_selectivity_2020}\cite{meyes_ablation_2019}. This suggests that researchers should focus on properties that exist across multiple neurons in high-dimensional distributed representations. In this work, we provide the first quantifiable metric based on multivariate information theory for analysing DNNs and multi-neuron importance. 

Recent work has used mutual information (MI) for individual neuron importance \cite{liu_understanding_2018}. However, the MI can identify the non-linear dependencies only between pairs of variables. Considering higher-order interactions of more than three variables allows a quantification of the informational content in terms of synergy and redundancy. Synergy arises from statistical interactions that can be found collectively in the whole (i.e. emergence) but not in the parts when considered in isolation, while redundancy refers to the information shared across variables \cite{timme_synergy_2014}. In this paper, we argue that the generalization abilities of DNNs is due to the networks ability to make a decision that emerges from the higher-order interactions of multiple synergistic neurons (i.e., the whole). In this work, we use a multivariate information theory measure called the O-information that scales to higher-order interactions\cite{rosas_quantifying_2019} to study the properties of multiple neurons in artificial neural networks. 

Current methods verify importance by removing important neurons from the network and evaluating the drop in classification performance. This approach results in a new model trained on data from a different training distribution. As a result, the test data is different from what has been seen from training. Without re-training, it is difficult to determine if the drop in accuracy is due to the distributional shift or if those neurons are actually important. To avoid this issue, we focus on the lottery ticket hypothesis (LTH)\cite{frankle_lottery_2018}. The LTH claims each network contains a sub-network such that when trained in isolation, it matches or exceeds the performance of the original network. If the synergistic neurons are important for classification then re-training these neurons should result in a minimal drop in accuracy. Our contributions are as following : 
\begin{enumerate}
    \item We present, to the best of our knowledge, the first quantifiable metric for multi-neuron feature importance
    \item We show that the features in the early layers of a DNN are redundant while the last layers contain synergistic features suggesting that the network specializes as the depth increase across the layers
    \item We provide an alternative perspective to verify importance by showing that re-training a sub-network based on the synergistic neurons results in a minimal drop in performance
\end{enumerate}
\textbf{Impact} Knowing which groups of neurons are important can: 
\begin{itemize}
\item Provide interpretability of the internal decisions in a network to end users\cite{li_interpretable_2021}
\item Explain generalization of DNNs\cite{kawaguchi_generalization_2017}
\item Enable compression of networks using pruning\cite{liang_pruning_2021} and the lottery ticket hypothesis\cite{frankle_lottery_2018}
\end{itemize}
\section{Related Work}

\textbf{Pairwise Mutual Information} MI is commonly used in deep learning as a loss function for training, quantifying importance of individual neurons\cite{liu_understanding_2018}, provide generalization bounds\cite{asadi_chaining_2018}. The limitation of all these approaches is that they rely on pairwise interactions, ignoring the information present in groups of neurons. 

\textbf{Multivariate Mutual Information} A few studies focused on multivariate information theory for analysing neural networks but they rely on the partial information decomposition which is difficult to scale when using more than three variables\cite{yu_understanding_2020}\cite{tax_partial_2017}. The information bottleneck \cite{tishby_deep_2015} uses multivariate MI but it considers the largest subset of neurons based on the entire layer.
\section{Methodology}
In this work, we use the O-information to identify which combination of $n$ neurons $x_{l}^{1}, .., x_{l}^{n}$ in each hidden layer $l$ is synergistic or redundant for classification of the input images into labels $y$. Importantly, we consider $x_{l}^{n}$ to be a vector of features associated with a neuron in a layer $l$ before applying the activation function to avoid non-singular values resulting from a ReLU activation function. $y$ is a vector of the class labels corresponding to each input image. Both $x_{l}^{n}$ and $y$ are evaluated based on the test set. 
\subsection{Summary of models and datasets}
We use a fully connected neural network architecture consisting of three fully connected hidden layers : $fc_{1}$, $fc_{2}$, $fc_{3}$ and the output layer $fc_{out}$ with 10 neurons. Each hidden layer consists of 20 neurons and uses ReLU as its activation function, except for the last layer that uses a softmax function. We train 20 networks on MNIST for 50 epochs, each reaching at least 96 percent accuracy on the training and test set. Training was performed using stochastic gradient descent with a learning rate of 0.01 and a batch size of 32. The training set consists of 60000 images while the test set consists of 10000 images.  We did not include any regularization techniques. For each experiment, we use 20 networks trained with a different random seed. 

\subsection{O-information}
Given a collection of $n$ random variables  $\bold{Z} = \{X_{1}, .., X_{n}\}$, we compute the O-information as \cite{rosas_quantifying_2019}:
\begin{equation}
\Omega_{n}(\bold{Z}) = (n - 2)H(\bold{Z}) + \sum_{j=1}^{n} [H(Z_{j}) - H(\bold{Z} \backslash Z_{j})]
\label{eq:1}
\end{equation}
where $H$ is the entropy, and $\bold{Z} \backslash Z_{j}$ is the complement of $Z_{j}$ with respect to $\bold{Z}$. If $\Omega_{n}(\bold{Z}) > 0$ the system is redundancy-dominated, while if $\Omega_{n}(\bold{Z}) < 0$  it is synergy-dominated. For a detailed description of the O-information, we refer to \cite{rosas_quantifying_2019}. 
\subsection{Layerwise multi-neuron importance using the O-information}
In order to use the O-information for feature importance, we need a way to construct random variables based on the features $x_{l}^{n}$ and target label $y$ that will be used as input to estimate the joint entropy in equation \ref{eq:1}. Most approaches using information-theoretic quantities rely on binning to transform the continuous samples into a discrete value. In this way the discrete entropy can be estimated in closed-form. However, binning relies on parameter tuning and is therefore vulnerable to bias. It also relies on a large amount of samples to reach an accurate estimate. To avoid these issues, we use the Gaussian Copula approach described in \cite{ince_statistical_2017}. In summary, this method transforms the marginals to standard Gaussian variables, to which we can apply the computationally efficient closed-form solution of the parametric joint entropy. 

\section{Experiments}
\subsection{Layerwise synergy and redundancy using the O-information}
In this experiment, we use the O-information to compare for each layer $l$ the optimal synergy and redundancy. We calculate the O-information $\Omega_{k}(x_{l}^{1}, .., x_{l}^{k}, y)$ for multiplets (i.e. subsets of features associated with $k$ neurons $x_{l}^{1}, .., x_{l}^{k}$ and the labels $y$) where $k$ is between 2 and the total number of neurons in the layer $n$. We calculate the synergy by minimizing $\Omega_{k}$ for each multiplet. Likewise, we calculate the redundancy by maximizing $\Omega_{k}$ for each multiplet. As an exhaustive search of each combination is computationally infeasible for large k, we propose a greedy search strategy by using an exhaustive search for $k = 6$, and larger $k$ values are handled by adding one variable at a time to the best multiplet of $k-1$ variables. 
\begin{figure}[!htb]
\minipage{0.32\textwidth}
  \includegraphics[width=\linewidth]{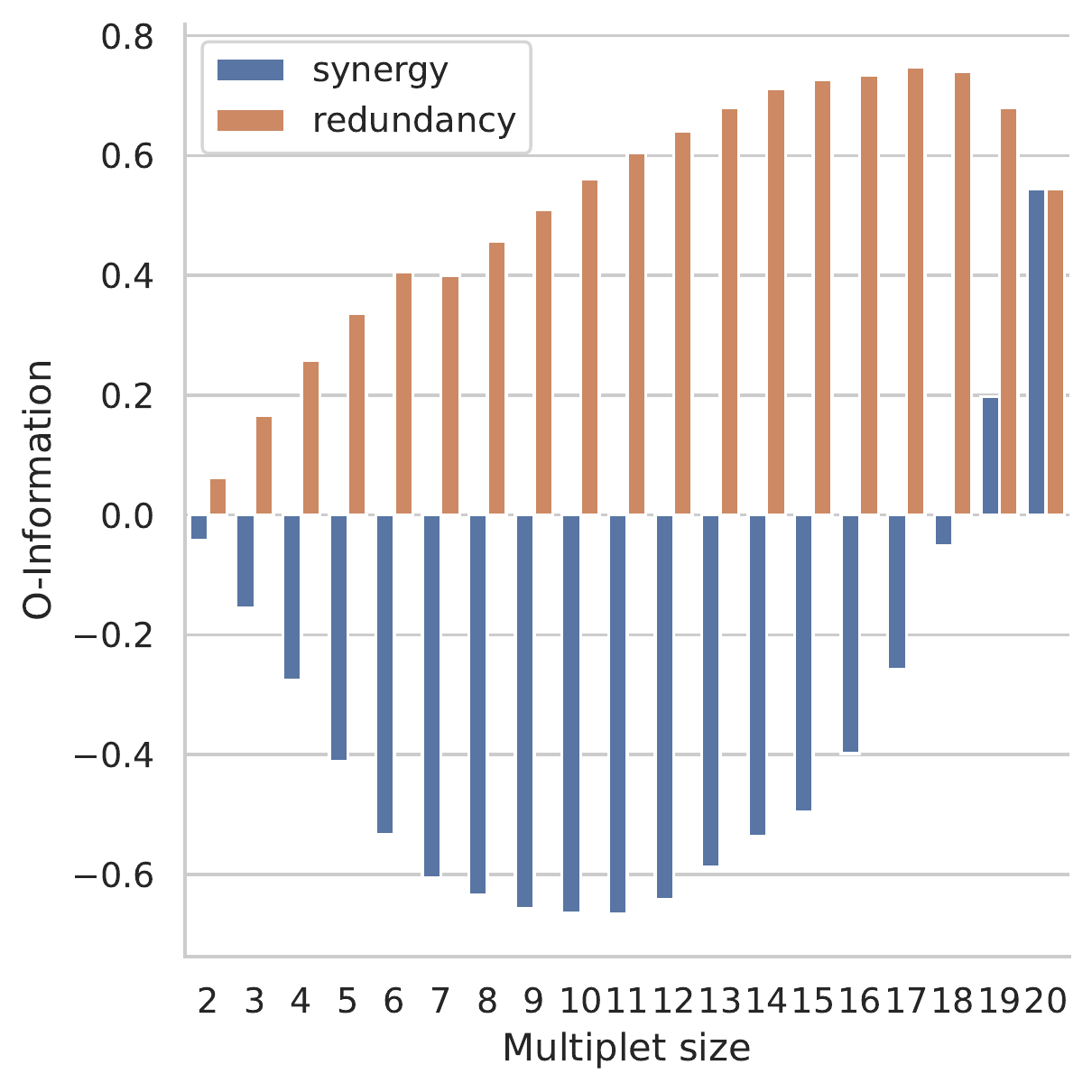}
  \captionsetup{labelformat=empty}
  \caption{(a) First hidden layer}
  \nocaption
\endminipage\hfill
\minipage{0.32\textwidth}
   \includegraphics[width=\linewidth]{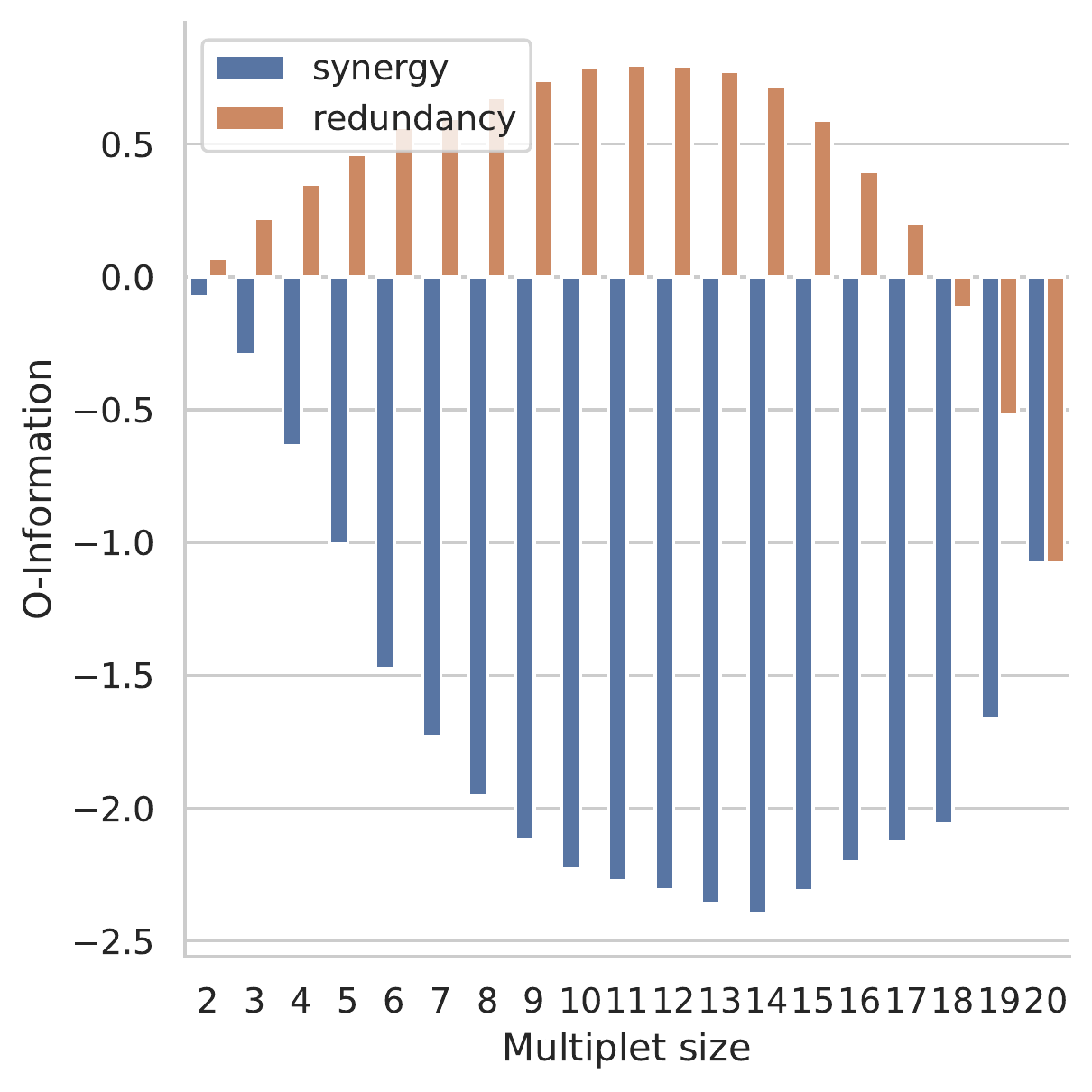}
  \captionsetup{labelformat=empty}
  \caption{(b) Second hidden layer}
  \nocaption
\endminipage\hfill
\minipage{0.32\textwidth}%
   \includegraphics[width=\linewidth]{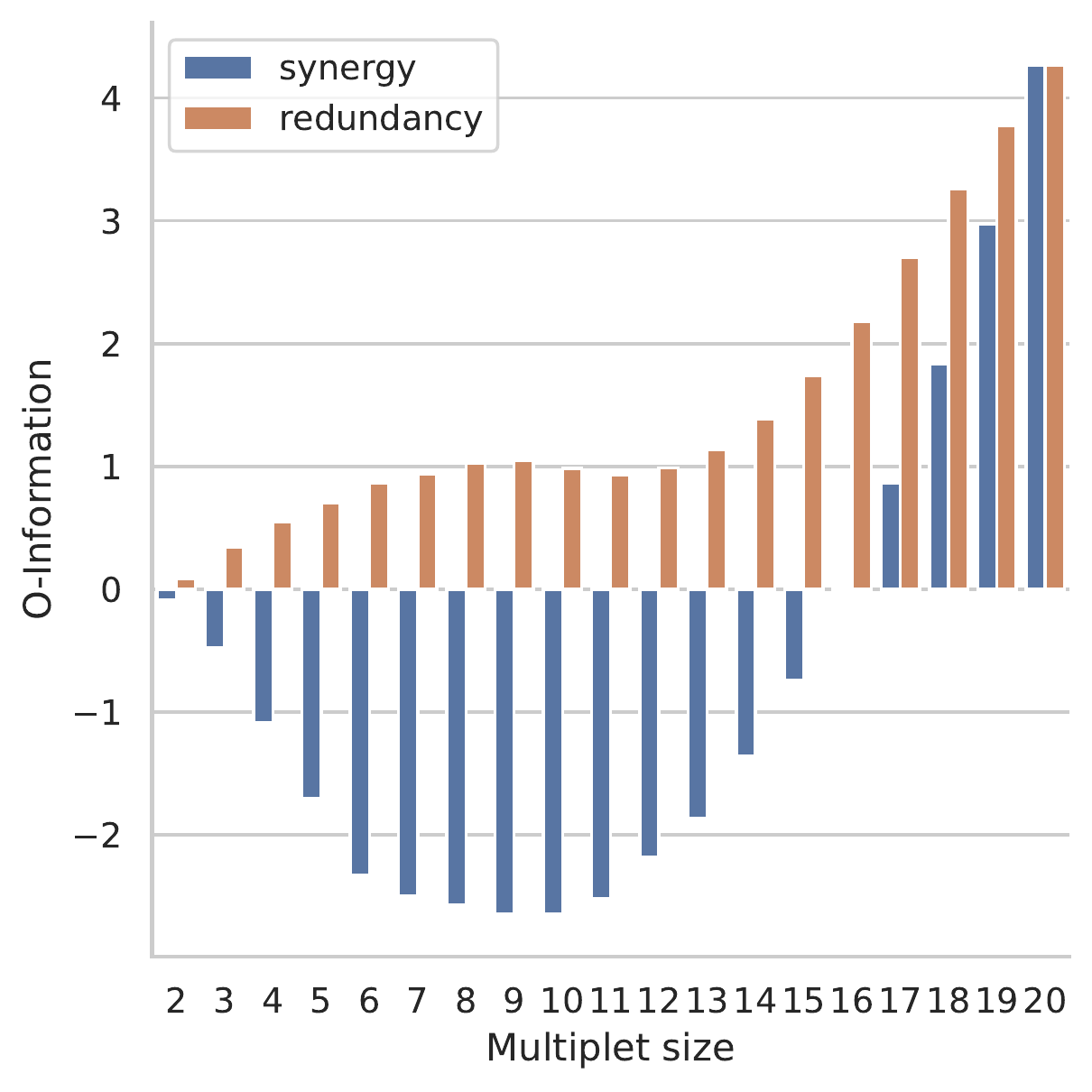}
  \captionsetup{labelformat=empty}
  \caption{(c) Third hidden layer}
  \nocaption
\endminipage
\caption{Layerwise bar plots of the synergy and redundancy in each layer for multiplets of increasing size. The experiments are averaged over 20 neural networks trained with a different random seed.}
\label{fig:exp1}

\end{figure}

As can be seen in figure \ref{fig:exp1}, the initial layers are dominated by redundant information which seems to suggest that the features in the first layer extract local features (i.e. edges) that are commonly shared among the classes. On the other hand, the last layer contains mainly synergistic information which can be explained by the network specializing for the classification decision. Our findings seem to contradict the results in \cite{liu_understanding_2018}, claiming that the last layers contain redundant information. We believe that these results empirically validates that pairwise MI cannot efficiently measure the synergy and redundancy of a multivariate systems. 

\subsection{Evaluating importance through re-training based on the lottery ticket hypothesis}
In this experiment, we empirically validate our hypothesis that the synergistic neurons are important for generalization by re-training the network with initial weights based on the these neurons. The remaining neurons are frozen during training. To achieve a fair comparison, we compare our method against three baselines. The first baseline randomly removes the neurons from the network. The second and third is based on the Gaussian Copula formulation of the mutual information (MI) between a single neuron and a discrete target\cite{ince_statistical_2017}. To take into account the discrete target, the former uses a mixture distribution, while the latter is based on ANOVA. As in \cite{ince_statistical_2017}, we perform cumulative ablation by ranking the individual neurons according to the highest MI.
\begin{figure}[!htb]
\includegraphics[scale=0.3]{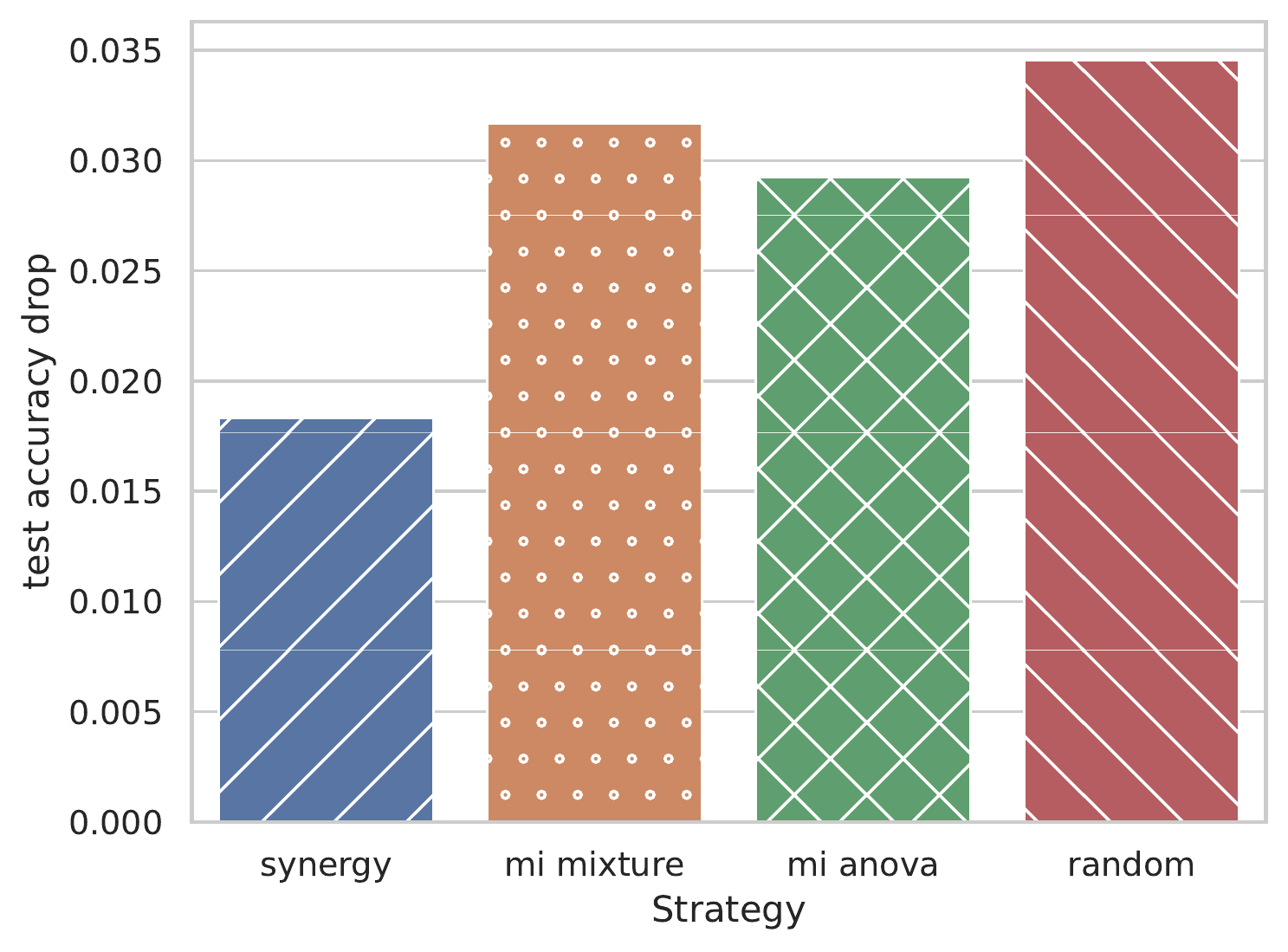}
\centering
\caption{Evaluating importance by training the re-initialized weights of a subnetwork based on the best synergistic multiplets in each layer}
\label{fig:exp2}
\end{figure}

As can be seen in figure \ref{fig:exp2}, re-training the synergistic neurons results in lowest accuracy drop. This seems to suggest that our method is able to identify the important neurons in a network. On the other hand, both MI methods have performances similar to the random strategy. If we can scale our method to provide accurate estimation of the synergy, we believe it can be used to identify optimal sub-networks in a single iteration. This is in contrast with current pruning methods for the LTH which rely on multiple iterations to remove a fixed number of neurons. 
\section{Limitations}
One of the main limitations of this work is the relatively small network and dataset. To increase the diversity of the experiments, we plan on extending this work to larger fully connected networks of 100 neurons on fashionMNIST\cite{xiao_fashion-mnist_2017} and VGG16\cite{simonyan_very_2014} on CIFAR10\cite{krizhevsky_learning_2009}. This raises two issues. First, we need a way to scale the greedy search to larger multiplets without sacrificing the accuracy of the estimation. Second, we have to use convolutions which require reasoning about feature maps containing many neurons rather then just one. A naive solution for the latter might be to average the feature maps into a scalar value but this ignores the rich information contained in the combinations of neurons. To address both issues, we plan on using evolutionary search strategies based on Jax to enable efficient parallel optimization of the multiplets\cite{tang_evojax_2022}. Another solution we are considering is to decompose the network into modular groups of neurons based on the weights using spectral clustering as in \cite{hod_detecting_2021}. 
\section{Conclusion}
In this work, to the best of our knowledge, we proposed the first method for quantifying multi-neuron feature importance\cite{matthew_l_leavitt_towards_2020}. We verified importance by re-training a sub-network based on the best synergistic neurons. In future work, we aim to explore the contribution of the features with respect to the individual classes. Another avenue for future research is to consider the training dynamics of DNNs using the information bottleneck(IB) principle\cite{tishby_deep_2015}. The IB considers the entire layer as input which compared to our method ignores the patterns of the intermediate combinations of neurons. From a practical perspective, we believe that the synergistic neurons can be used as a metric for pruning and as an optimization objective for unsupervised representation learning \cite{vera_role_2018}.

\newpage
\bibliography{references2}

\begin{thebibliography}{22}
\providecommand{\natexlab}[1]{#1}
\providecommand{\url}[1]{\texttt{#1}}
\expandafter\ifx\csname urlstyle\endcsname\relax
  \providecommand{\doi}[1]{doi: #1}\else
  \providecommand{\doi}{doi: \begingroup \urlstyle{rm}\Url}\fi

\bibitem[Asadi et~al.()Asadi, Abbe, and Verdú]{asadi_chaining_2018}
A.~Asadi, E.~Abbe, and S.~Verdú.
\newblock Chaining mutual information and tightening generalization bounds.
\newblock 31.

\bibitem[Frankle and Carbin()]{frankle_lottery_2018}
J.~Frankle and M.~Carbin.
\newblock The lottery ticket hypothesis: Finding sparse, trainable neural
  networks.

\bibitem[Hod et~al.()Hod, Casper, Filan, Wild, Critch, and
  Russell]{hod_detecting_2021}
S.~Hod, S.~Casper, D.~Filan, C.~Wild, A.~Critch, and S.~Russell.
\newblock Detecting modularity in deep neural networks.

\bibitem[Ince et~al.()Ince, Giordano, Kayser, Rousselet, Gross, and
  Schyns]{ince_statistical_2017}
R.~A. Ince, B.~L. Giordano, C.~Kayser, G.~A. Rousselet, J.~Gross, and P.~G.
  Schyns.
\newblock A statistical framework for neuroimaging data analysis based on
  mutual information estimated via a gaussian copula.
\newblock 38\penalty0 (3):\penalty0 1541--1573.
\newblock \doi{10.1002/hbm.23471}.
\newblock Publisher: Wiley Online Library.

\bibitem[Kawaguchi et~al.()Kawaguchi, Kaelbling, and
  Bengio]{kawaguchi_generalization_2017}
K.~Kawaguchi, L.~P. Kaelbling, and Y.~Bengio.
\newblock Generalization in deep learning.

\bibitem[Krizhevsky and Hinton()]{krizhevsky_learning_2009}
A.~Krizhevsky and G.~Hinton.
\newblock Learning multiple layers of features from tiny images.
\newblock Publisher: Citeseer.

\bibitem[Li et~al.()Li, Xiong, Li, Wu, Zhang, Liu, Bian, and
  Dou]{li_interpretable_2021}
X.~Li, H.~Xiong, X.~Li, X.~Wu, X.~Zhang, J.~Liu, J.~Bian, and D.~Dou.
\newblock Interpretable deep learning: Interpretation, interpretability,
  trustworthiness, and beyond.
\newblock URL \url{http://arxiv.org/abs/2103.10689}.

\bibitem[Liang et~al.()Liang, Glossner, Wang, Shi, and
  Zhang]{liang_pruning_2021}
T.~Liang, J.~Glossner, L.~Wang, S.~Shi, and X.~Zhang.
\newblock Pruning and quantization for deep neural network acceleration: A
  survey.
\newblock 461:\penalty0 370--403.
\newblock Publisher: Elsevier.

\bibitem[Liu et~al.()Liu, Amjad, and Geiger]{liu_understanding_2018}
K.~Liu, R.~A. Amjad, and B.~Geiger.
\newblock Understanding individual neuron importance using information theory.

\bibitem[{Matthew L. Leavitt} et~al.({\natexlab{a}}){Matthew L. Leavitt},
  Leavitt, {Matthew L. Leavitt}, and
  Morcos]{matthew_l_leavitt_selectivity_2020}
{Matthew L. Leavitt}, M.~Leavitt, {Matthew L. Leavitt}, and A.~S. Morcos.
\newblock Selectivity considered harmful: evaluating the causal impact of class
  selectivity in {DNNs}.
\newblock {\natexlab{a}}.
\newblock {MAG} {ID}: 3009079798.

\bibitem[{Matthew L. Leavitt} et~al.({\natexlab{b}}){Matthew L. Leavitt},
  Leavitt, {Matthew L. Leavitt}, and Morcos]{matthew_l_leavitt_towards_2020}
{Matthew L. Leavitt}, M.~Leavitt, {Matthew L. Leavitt}, and A.~S. Morcos.
\newblock Towards falsifiable interpretability research.
\newblock {\natexlab{b}}.
\newblock {MAG} {ID}: 3094509718.

\bibitem[Meyes et~al.()Meyes, Lu, de~Puiseau, and Meisen]{meyes_ablation_2019}
R.~Meyes, M.~Lu, C.~W. de~Puiseau, and T.~Meisen.
\newblock Ablation studies in artificial neural networks.

\bibitem[Morcos et~al.()Morcos, Barrett, Rabinowitz, and
  Botvinick]{morcos_importance_2018}
A.~S. Morcos, D.~G. Barrett, N.~C. Rabinowitz, and M.~Botvinick.
\newblock On the importance of single directions for generalization.

\bibitem[Rosas et~al.()Rosas, Mediano, Gastpar, and
  Jensen]{rosas_quantifying_2019}
F.~E. Rosas, P.~A. Mediano, M.~Gastpar, and H.~J. Jensen.
\newblock Quantifying high-order interdependencies via multivariate extensions
  of the mutual information.
\newblock 100\penalty0 (3):\penalty0 032305.
\newblock \doi{10.1103/PhysRevE.100.032305}.
\newblock Publisher: {APS}.

\bibitem[Simonyan and Zisserman()]{simonyan_very_2014}
K.~Simonyan and A.~Zisserman.
\newblock Very deep convolutional networks for large-scale image recognition.

\bibitem[Tang et~al.()Tang, Tian, and Ha]{tang_evojax_2022}
Y.~Tang, Y.~Tian, and D.~Ha.
\newblock {EvoJAX}: Hardware-accelerated neuroevolution.

\bibitem[Tax et~al.()Tax, Mediano, and Shanahan]{tax_partial_2017}
T.~Tax, P.~A. Mediano, and M.~Shanahan.
\newblock The partial information decomposition of generative neural network
  models.
\newblock 19\penalty0 (9):\penalty0 474.
\newblock \doi{10.3390/e19090474}.
\newblock Publisher: Multidisciplinary Digital Publishing Institute.

\bibitem[Timme et~al.()Timme, Alford, Flecker, and Beggs]{timme_synergy_2014}
N.~Timme, W.~Alford, B.~Flecker, and J.~M. Beggs.
\newblock Synergy, redundancy, and multivariate information measures: an
  experimentalist’s perspective.
\newblock 36\penalty0 (2):\penalty0 119--140.
\newblock \doi{10.1007/s10827-013-0458-4}.
\newblock Publisher: Springer.

\bibitem[Tishby and Zaslavsky()]{tishby_deep_2015}
N.~Tishby and N.~Zaslavsky.
\newblock Deep learning and the information bottleneck principle.
\newblock In \emph{2015 ieee information theory workshop (itw)}, pages 1--5.
  {IEEE}.
\newblock \doi{10.1109/ITW.2015.7133169}.

\bibitem[Vera et~al.()Vera, Piantanida, and Vega]{vera_role_2018}
M.~Vera, P.~Piantanida, and L.~R. Vega.
\newblock The role of the information bottleneck in representation learning.
\newblock In \emph{2018 {IEEE} International Symposium on Information Theory
  ({ISIT})}, pages 1580--1584. {IEEE}.

\bibitem[Xiao et~al.()Xiao, Rasul, and Vollgraf]{xiao_fashion-mnist_2017}
H.~Xiao, K.~Rasul, and R.~Vollgraf.
\newblock Fashion-mnist: a novel image dataset for benchmarking machine
  learning algorithms.

\bibitem[Yu et~al.()Yu, Wickstrøm, Jenssen, and
  Príncipe]{yu_understanding_2020}
S.~Yu, K.~Wickstrøm, R.~Jenssen, and J.~C. Príncipe.
\newblock Understanding convolutional neural networks with information theory:
  An initial exploration.
\newblock 32\penalty0 (1):\penalty0 435--442.
\newblock Publisher: {IEEE}.

\end{thebibliography}
\bibliographystyle{abbrvnat}

\end{document}